\documentclass[10pt,twocolumn,letterpaper]{article}

\usepackage[pagenumbers]{cvpr} %

\usepackage{graphicx}
\usepackage{amsmath}
\usepackage{amssymb}
\usepackage{booktabs}

\usepackage{times}
\usepackage{epsfig}
\usepackage{graphicx}
\usepackage{amsmath}
\usepackage{amssymb}

\usepackage{booktabs}    %
\usepackage[font=small]{caption}
\usepackage{multirow}
\usepackage{arydshln}
\usepackage{capt-of}
\usepackage{algorithm}
\usepackage{listings}
\usepackage{tabularx}
\usepackage[table]{xcolor}
\usepackage{colortbl}
\usepackage{cuted}
\usepackage{adjustbox}
\usepackage{microtype}

\usepackage{afterpage}
\usepackage{placeins}
\definecolor{newlightblue}{RGB}{0,75,255}
\usepackage[pagebackref,breaklinks,colorlinks, urlcolor={newlightblue}, citecolor={newlightblue}]{hyperref}

\usepackage[capitalize]{cleveref}
\crefname{section}{Sec.}{Secs.}
\Crefname{section}{Section}{Sections}
\Crefname{table}{Table}{Tables}
\crefname{table}{Tab.}{Tabs.}

\DeclareMathOperator{\tr}{tr}

\newcommand{\mypar}[1]{\vspace{-2mm}\paragraph{#1}}

\newcommand{\bx}{\mathbf{x}}
\newcommand{\bv}{\mathbf{v}}
\newcommand{\ba}{\mathbf{a}}

\newcommand{\bs}{\mathbf{s}}
\newcommand{\mcl}{\mathcal{L}}

\definecolor{lightyellow}{RGB}{255,255,170}

\def\cvprPaperID{715} %
\def\confName{CVPR}
\def\confYear{2022}

\begin{document}

\title{Mix and Localize: Localizing Sound Sources in Mixtures}

\author{ Xixi Hu\textsuperscript{1,2*} 
\and Ziyang Chen\textsuperscript{1*}
\and Andrew Owens\textsuperscript{1} 
\vspace{-0.5em}
\\ 
\and University of Michigan\textsuperscript{1} 
\hspace{2em} The University of Texas at Austin\textsuperscript{2}
}

\maketitle

{\let\thefootnote\relax\footnotetext{{\textsuperscript{*} Indicates equal contribution.}}}

\begin{abstract}
We present a method for simultaneously localizing multiple sound sources within a visual scene. This task requires a model to both group a sound mixture into individual sources, and to associate them with a visual signal. Our method jointly solves both tasks at once, using a formulation inspired by the contrastive random walk of Jabri et al. We create a graph in which images and separated sounds correspond to nodes, and train a random walker to transition between nodes from different modalities with high return probability. The transition probabilities for this walk are determined by an audio-visual similarity metric that is learned by our model. We show through experiments with musical instruments and human speech that our model can successfully localize multiple sounds, outperforming other self-supervised methods. Project site: \small{\url{https://hxixixh.github.io/mix-and-localize}}.

\end{abstract}
\vspace{-0.3em}

\vspace{-2mm} \section{Introduction}

Humans have the remarkable ability to localize many sounds at once~\cite{hawley1999speech}. Existing audio-visual sound localization methods, by contrast, are generally trained with the assumption that only a single sound source is present at a time, and largely lack mechanisms for grouping the contents of a scene into multiple audio-visual events. 

This problem is often addressed through contrastive learning~\cite{senocak2018learning,arandjelovic2018objects,owens2018audio,afouras2020self,arandjelovic2017look}. These methods generally produce a single embedding for the audio, representing the sound source, and an embedding for each image patch, representing the sound's possible locations. They then learn cross-modal correspondences, such that image patches and sounds that co-occur within the same scene are brought close together, and pairings that do not co-occur are pulled apart. %
Extending this approach to multiple sound sources seemingly requires solving two different problems: separating the sources from a sound mixture, and localizing them within an image.

We propose a simple model that jointly addresses both of these problems. Our model uses cycle consistency to group a scene into sound sources, inspired by the contrastive random walk~\cite{jabri2020space}. It produces multiple embedding vectors for a sound mixture, each representing a different sound source, and an audio-visual similarity metric that associates them with their corresponding image content. %
This similarity metric defines the transition probabilities for a random walk on a graph whose nodes correspond to images and predicted sound sources. Our model performs a random walk that transitions from the audio to images, and then back. We learn a similarity metric that maximizes the probability of {\em cycle-consistency} (\ie, return probability) for the walk. After training, we create an attention map between an image and each sound source by estimating the similarity score between the audio and visual embeddings.

\begin{figure}[t]
    \centering
    \includegraphics[width=1\columnwidth]{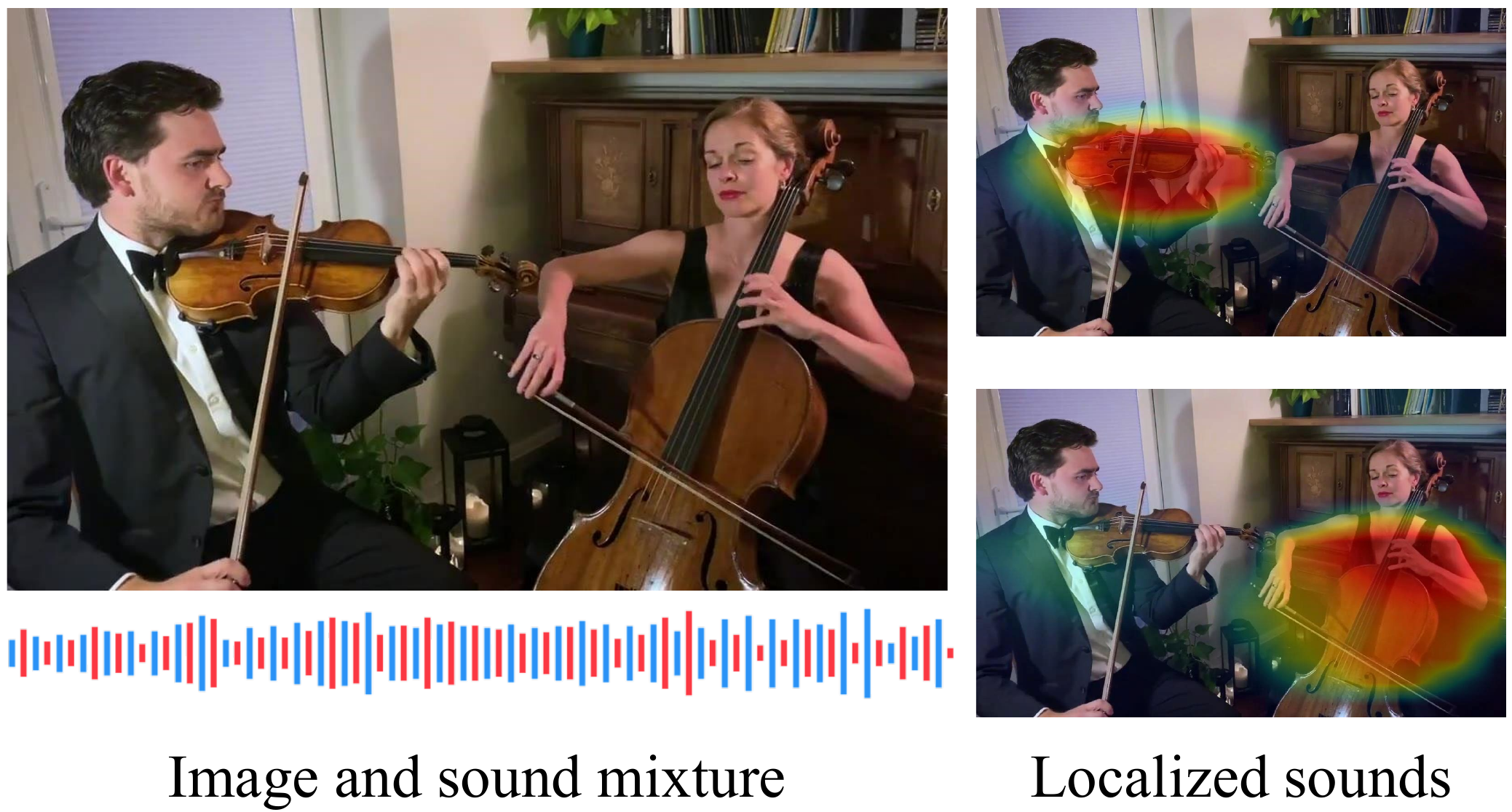}
    \caption{{\bf Cycle-consistent multi-source localization.} 
    Our model jointly learns to separate a sound mixture into sound sources, and to localize these sources within an image. To do this, we use a self-supervised grouping method based on cycle-consistent random walks. We show sound source localization results from our model. } \vspace{-5mm}
    \label{fig:teaser}
\end{figure}

Obtaining a cycle-consistent walk requires extracting sound sources from the mixture, and associating each one with distinct image content. Consequently, this formulation has several advantages over other self-supervised audio-visual localization methods. It separates sounds from a mixture, and explicitly groups the scene into discrete sound-making objects. The model is also simple, and can be implemented as a straightforward extension of previous contrastive methods.

We evaluate our model on synthetic and real-world sound mixtures containing  musical instruments~\cite{zhao2018sound,chen2020vggsound} and human speakers~\cite{Chung18b}. We find that, in comparison to other self-supervised localization methods, our model is more accurate in localizing sounds in multi-source mixtures.

\section{Related Work}

\vspace{3mm}
\mypar{Sound source separation.}
There has been a long history of methods for separating monaural sound mixtures. Early work addressed this problem with probabilistic models~\cite{cichocki2009nonnegative, roweis2000one} and recent work has tackled it via deep neural networks~\cite{hershey2016deep,yu2017permutation}. These often use a ``mix-and-separate"~\cite{zhao2018sound} training procedure~\cite{stoller2018wave, jansson2017singing}. Other work combines source separation with visual cues. Zhao~\etal~\cite{zhao2018sound} proposed to separate different musical instruments by associating separated audio sources with pixels in the video, and later used optical flow~\cite{zhao2019sound} to provide motion cues. Gao \etal~\cite{gao2021visualvoice} jointly solve audio-visual speech separation with a multi-task learning framework by incorporating cross-modal face-voice attributes and lip motion. Tian \etal~\cite{tian2021cyclic} jointly learn sound separation and sounding object visual grounding, using an approach they term cyclic co-learning. Chatterjee \etal~\cite{chatterjee2021visual} model visual signals into scene graphs and learn to separate sounds by co-segmenting subgraphs and associated audio. Majumder \etal~\cite{majumder2021move2hear} introduce the active audio-visual source separation task that an agent learns movement policies to improve sound separation qualities.
Like these works, we jointly localize and separate sound. However, we do not generate separated audio: we obtain embeddings that represent the separated sound sources using contrastive learning.

\mypar{Audio-visual sound source localization.}
The co-occurrence of audio and visual cues in videos has been leveraged for sound source localization~\cite{senocak2018learning,harwath2018jointly,tian2018audio,qian2020multiple,hu2019deep}. Researchers exploit audio-visual correspondence by matching audio and visual signals from the same video.
Arandjelovi\'{c} and Zisserman~\cite{arandjelovic2018objects, arandjelovic2017look} measure the similarity between learned image region and audio representations and use multi-instance learning to localize sound sources. 
Owens and Efros~\cite{owens2018audio} use class activation maps~\cite{zhou2016learning} to visualize the area contributing to solving audio-visual synchronization tasks. 
Chen et al.~\cite{chen2021localizing} mine hard negative image locations in cross-modal contrastive learning to obtain better sound localization results.
Hu \etal~\cite{hu2020discriminative} extend \cite{arandjelovic2018objects} and use clustering to generate pseudo-class labels, achieving class-aware sound source localization with mixed sounds. While we have a similar goal, our approach is {entirely unsupervised}, and does not require semi-supervised learning with labels, either at training or test time. %
Our work is motivated by them and aims to localize different sound sources in multi-source sound mixtures.

\mypar{Audio-visual self-supervision.} 
Apart from sound source localization and separation, many recent works have proposed to use paired audio-visual data for representation learning and other tasks as well. 
Owens \etal~\cite{owens2016visually} learned visual representations for materials from impact sounds. Other work has learned features, scene structure, and geometric properties from sounds~\cite{owens2016ambient, chen2021structure, gao2020visualechoes}, or learns multisensory representations for both audio and vision~\cite{owens2018audio, korbar2018cooperative, xiao2020audiovisual}. Asano \etal~\cite{asano2020labelling} proposed self-supervised clustering and representation learning approach for providing labels to multimodal data.
Other work has learned active speaker detection~\cite{chung2016out, afouras2020self}, up-mixing the mono audio~\cite{gao20192, yang2020telling, rachavarapu2021localize}, cross-modal distillation~\cite{aytar2016soundnet, valverde2021there}.
We take inspiration from them and learn the representation of mixture sounds.

\mypar{Graph-based representation learning.}
A number of recent works use graphs to learn image and video segmentation~\cite{shi2000normalized} and space-time correspondence~\cite{vondrick2018tracking,wang2019learning,jabri2020space}. Jabri \etal~\cite{jabri2020space} propose to use graphs to learn the visual correspondence between several frames clipped from video, and the graph is constructed by connecting patches in spatio-temporal neighborhoods. Bian \etal~\cite{bian2022learning} propose multi-scale contrastive random walks to obtain pixel-level correspondence between frames. 
We extend this approach to a multi-modal learning domain, rather than to video representation learning.

\newcommand{\simfn}[0]{\phi}
\newcommand{\iemb}[0]{f}
\newcommand{\aemb}[0]{g}

\section{Method}
\begin{figure*}[t]
    \centering
    \includegraphics[width=1\textwidth]{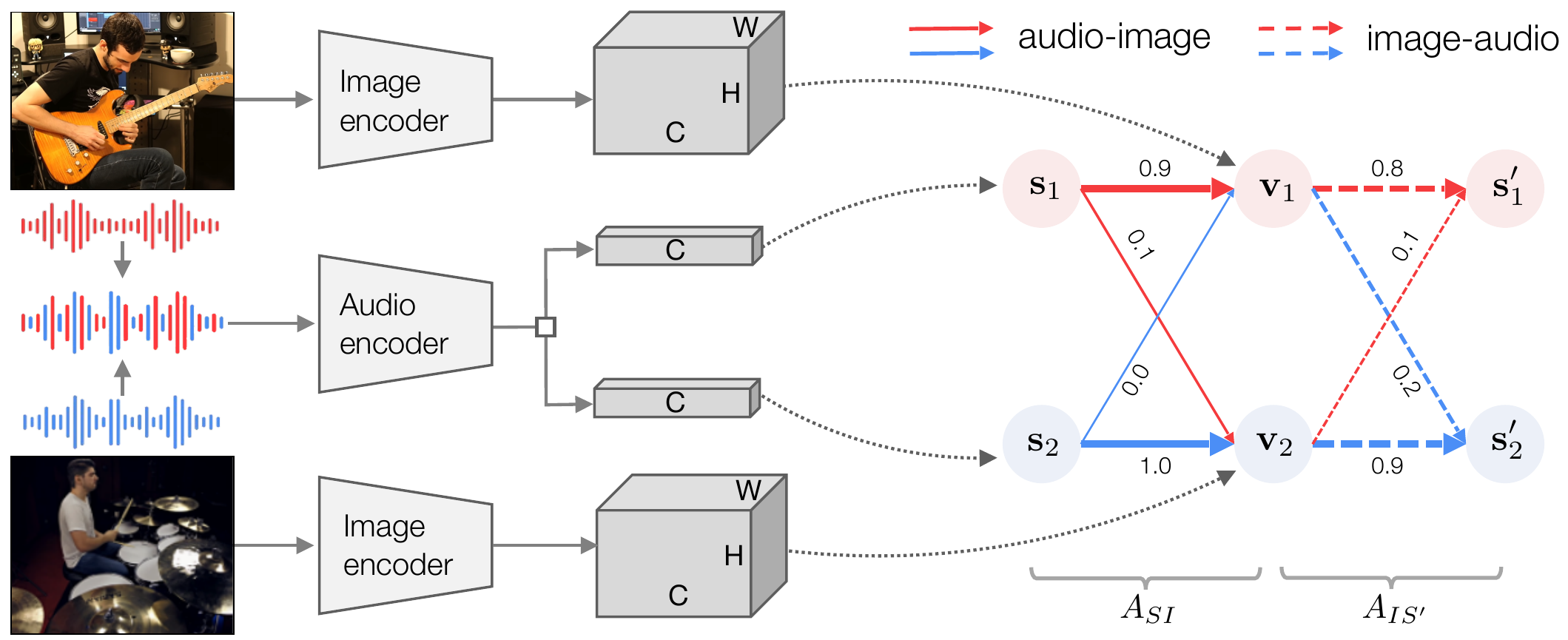} 
    \caption{{\bf Audio-visual random walks.} %
    We learn representations for separating and localizing sounds. We generate a synthetic  mixture by summing waveforms from multiple videos (we show $k=2$ videos). Our model estimates embedding vectors from the audio mixture, representing each sound source, and learns an audio-visual similarity metric that associates image regions with the extracted sources. We solve a cycle-consistency problem in a graph. Edges connect each audio node to nodes that represent each image. A random walker is trained to walk from each audio node to an image node, then back to the audio.  %
    Our model learns to guide the random walker to the node it began its walk (\ie to maximize its cycle consistency) using transition probabilities derived from the similarity function. 
    }
    \label{fig:method}
\end{figure*}

Our goal is to perform {\em multi-source} audio-visual sound localization. Given audio $\ba$ and corresponding image $\bv$, we will parse the scene into discrete sound sources and localize them within an image. We frame this as a representation learning problem. We produce embedding vectors $\bs_1, \bs_2, \cdots, \bs_k$ from $\ba$, representing the $k$ visible sound sources, and associate them with visual embeddings for image regions $\bx_1, \bx_2, \cdots, \bx_m$.

If we were given ground-truth correspondences between sources $\bs_i$ and image regions $\bx_i$, then contrastive localization methods~\cite{arandjelovic2018objects, arandjelovic2017look, senocak2018learning} could straightforwardly be applied to this problem. However, the sound sources are {\em latent} and must be estimated from the audio. We propose to jointly solve both problems: we produce audio embeddings from a mixture that provide cycle consistent matches with the image regions.

\mypar{Single-source localization as a random walk.} \label{sec:cross-mod-rand-walk}
As a preliminary step toward solving this problem, we start by considering the simpler single-source localization problem. As in previous  work~\cite{arandjelovic2018objects,senocak2018learning,afouras2020self}, we can address this problem using contrastive learning. We learn an embedding $\iemb(\bx) \in \mathbb{R}^C$ for each image region and another embedding $\aemb(\ba)$ for audio. In practice, we compute the image embeddings using a fully convolutional network that operates on full images. We define a cross-modal similarity metric:
\begin{equation}
\simfn(\bv, \ba) = \max_{\bx_i} \iemb(\bx_i)^\top\aemb(\ba),
\label{eq:mapscore}
\end{equation} 
where the pooling is performed over all image regions $\bx_i$ in $\bv$. Following~\cite{arandjelovic2018objects}, we summarize the similarity of the whole image by taking the $\max$ over all the image regions, under the assumption that the sound source occupies a small portion of the image. We can learn audio and visual representations using contrastive learning:  $\bv_i$ should be more similar to its corresponding audio track $\ba_i$ than to $n-1$ other audio examples. If ${A}_{IS}(i, j)$ is the similarity between $\bv_i$ and $\ba_j$, these similarities can be formulated as: 
\begin{equation}
\begin{aligned}
  {A}_{IS}(i, j) & = \frac{\exp(\simfn(\bv_i, \ba_j)/ \tau) }{\sum_{k=1}^n \exp(\simfn(\bv_i, \ba_k)/ \tau)}, 
  \label{eq:AIS}
\end{aligned}
\end{equation}
\noindent where $\tau$ is a temperature hyperparameter~\cite{wu2018unsupervised}. The summation in the denominator iterates over both $\ba_j$ and the $n-1$ other audio examples, and ${A}_{IS} \in \mathbb{R}^{n \times n}$. We maximize the diagonal of $A_{IS}$ using the InfoNCE loss~\cite{oord2018representation}:
\begin{equation}
    \mathcal{L}_{\mathtt{corresp}} = -\frac{1}{n}\tr(\log(A_{IS})),
\end{equation}
where the $\log$ is performed elementwise. After training, the dot product between the image and audio embeddings $\iemb(\bx_i)^\top \aemb(\ba)$ can be interpreted as the likelihood of $\bx_i$ being the location of a sound source, since this conceptually represents the correlation between the visual and audio signals. 

One can interpret $A_{IS}$ as the transition matrix of a random walk that moves from images to sounds, and $\mathcal{L}_{\mathtt{corresp}}$ as a penalty for transitioning to an incorrect sound. One could also compute an analogous matrix $A_{SI}$ by matching from audio $\ba_i$ to the image  $\bv_j$ with softmax normalization (equivalent to normalizing columns in Eq.~\ref{eq:AIS}, rather than rows, and then transposing).

\begin{figure*}[t]
    \centering
    \includegraphics[width=1\textwidth]{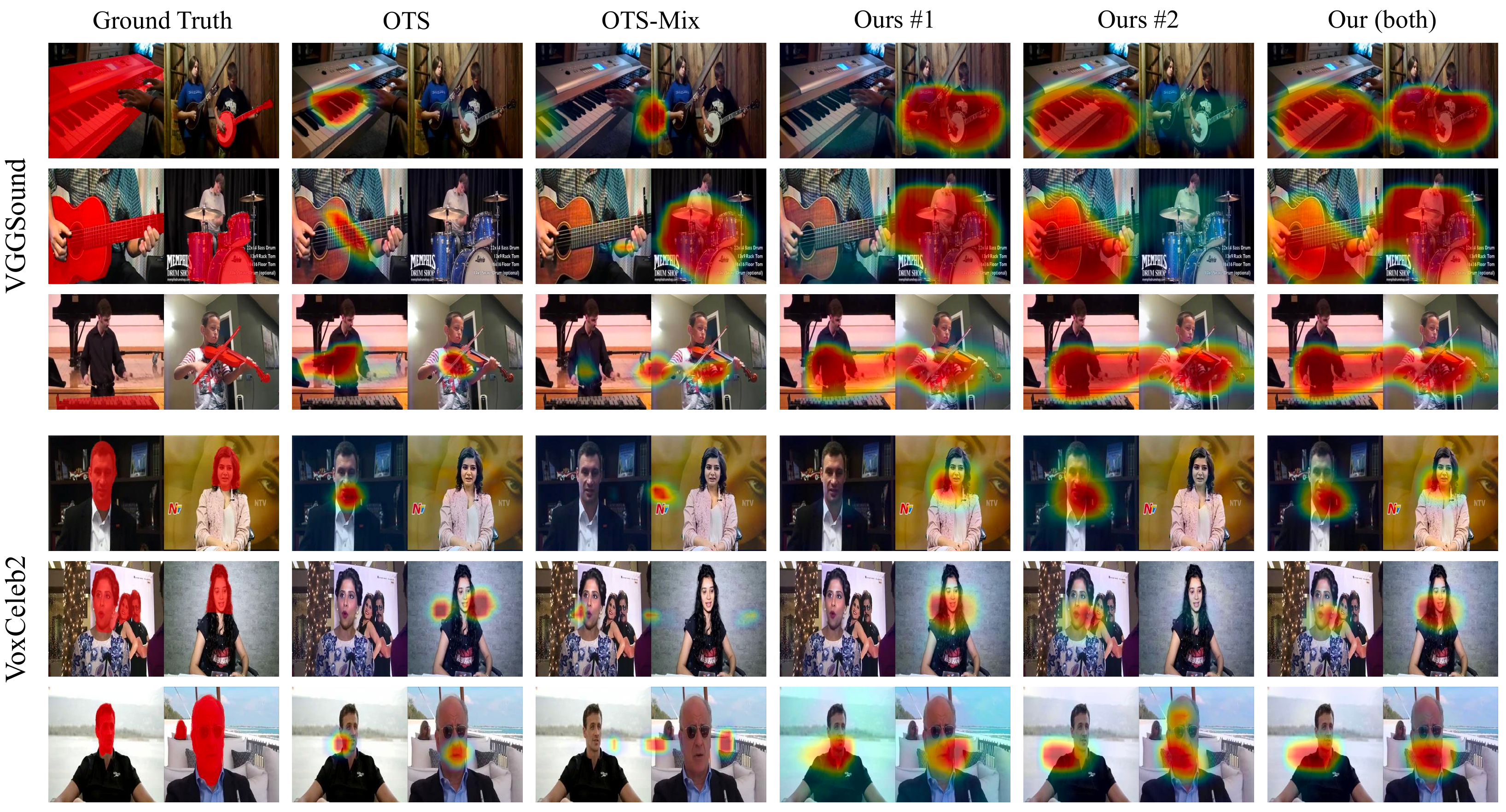}
    \caption{{{Multi-source localization results} on synthetic mixtures from the VGGSound-Instruments and VoxCeleb2 datasets. We provide a comparison of localization maps generated by different methods. We show the two localization maps generated by our model's two embeddings. The color of the image region indicates its localization score, with the red regions having higher scores. 
    We show failure cases in the last row of each dataset.}
    }
    \label{fig:vggsound_ssl}
\end{figure*}

\mypar{Random walk with cycle-consistency.} 
Now, suppose that we do not know the ground-truth correspondence between images and audio, but merely that there is an unknown, one-to-one relationship between the separated audio embeddings and images. We use a {\em cycle consistent} random walk to jointly learn the audio embeddings and associate them with images.

We are given a synthetic sound mixture containing $k$ components, created by summing $k$ different waveforms, along with the corresponding $k$ images they were taken from (we use $k=2$ in our experiments). We construct a directed graph containing nodes for each sound source and each image. Its edges lead from sound sources to images and back (Figure~\ref{fig:method}), with transition probabilities determined by audio-visual similarity.  A random walker in this graph starts from an audio node $\bs_{i}$, travels to an image node $\bv_{j}$ and arrives at another audio node $\bs_{t}$. 

Inspired by recent works on visual correspondence~\cite{wang2019learning,jabri2020space}, we use a {\em cycle-consistency} loss to guide the random walker. 
While we do not know whether a given $\bs_i$ and $\bv_j$ pair belong to the same audio event, we do know that there should be a one-to-one relationship between images and sounds. %
As in Eq.~\ref{eq:AIS}, we compute a matrix $A_{IS} \in \mathbb{R}^{k \times k}$ such that $A_{IS}(i, j)$ measures the similarity between image embedding $\bv_i$ and audio embedding $\bs_j$. We encourage audio to return to itself with high probability in the random walk:
\begin{equation}
\begin{aligned}
     \mcl_{\mathtt{cyc}} = -\frac{1}{k}\tr(\log({A}_{SI} A_{IS})).
  \label{eq:loss_cyc}
\end{aligned}
\end{equation}
\noindent
Under this loss, the model is encouraged to maximize the probability of associating a sound to highly discriminative image regions---those that can successfully select this sound over all others. Since the embeddings are produced from a sound mixture, a natural choice is to represent its sound sources. After training, the dot products $\iemb(\bx_i)^\top \aemb(\bs_j)$ convey the location of each of the $k$ sound sources.

\mypar{Data augmentation.} We found that our models could often quickly drop the cycle consistency loss (Eq.~\ref{eq:loss_cyc}) to low values, since high-dimensional embedding vectors are often cycle-consistent by chance. We encourage the model to learn other useful invariances by using randomly-shifted versions of the audio when computing the transition matrix $A_{IS}$, resulting in a matrix we call $A_{IS'}$ (Fig.~\ref{fig:method}).

\section{Experiments}

\begin{table*}[!t]
\centering
\resizebox{\textwidth}{!}{
\begin{tabular}{c@{\hskip20pt}l@{\hskip20pt}ccccccccccc}

\toprule
\multicolumn{2}{c}{} & \multicolumn{3}{c}{\textbf{Single sound source}} & \multicolumn{8}{c}{\textbf{Multiple sound sources}} \\
\cmidrule{3-13}

\multicolumn{2}{c}{} & \multicolumn{3}{c}{MUSIC-Solo} & \multicolumn{4}{c}{MUSIC-Synthetic} & \multicolumn{4}{c}{MUSIC-Duet}\\
\cmidrule(lr){3-5} 
\cmidrule(lr){6-9}
\cmidrule(lr){10-13}
\multicolumn{2}{c}{}  & AP & AUC& IoU@0.5 & CAP & PIAP & AUC & CIoU@0.3 & CAP & PIAP & AUC & CIoU@0.3 \\

\midrule

\multirow{2}{*}{\shortstack[c]Semi-supervised} & Hu~\etal~\cite{hu2020discriminative} & - & 43.6 & 51.4 & - & - & 23.5 & 32.3 & - & - & 22.1 & 30.2 \\

    & Sound of Pixels + Matching~\cite{zhao2018sound, hu2020discriminative}              & -  & 43.3 & 40.5 & - & - & 11.8 & 8.1 & - & - & 16.8 & 16.8 \\
    \midrule
    \multirow{5}{*}{\shortstack[c]{Self-supervised}} & OTS~\cite{arandjelovic2018objects}                & \textbf{69.3} & 35.8 & 26.1 & 11.4 & 17.6 & 10.2 & 3.7 & 35.4 & 42.8 & 18.3 & 13.2 \\
    & OTS-mix~\cite{arandjelovic2018objects}            & 53.8 & 33.9 & 17.5 & 16.9 & 27.4 & 7.3 & 0 & 23.8 & 28.2 & 12.0 & 2.0  \\ 
    & Attention~\cite{senocak2018learning}              & - & 38.7 & \textbf{37.2} & - & - & 12.3 & 6.4 & - & - & 19.4 & 21.5 \\
    & DMC~\cite{hu2019deep}                             & - & 38.0 & 29.1 & - & - & 16.3 & 7.0 & - & - & 21.1 & 17.3 \\
    
\cdashlinelr{2-13} 
    & Ours                                              & 67.9 & \textbf{40.6} & 29.2 & \textbf{34.0}* & \textbf{39.7} & \textbf{20.2}* & \textbf{21.3}* & \textbf{47.4}* & \textbf{53.9} & \textbf{21.2}* & \textbf{26.3}*  \\
    
\bottomrule
\end{tabular}
}
\caption{Sound source localization performance on MUSIC dataset. Following previous work~\cite{hu2020discriminative, senocak2018learning}, IoU@0.5 measures ratio of successful samples at 0.5 threshold. Similarly, CIoU@0.3 measures the ratio of successful samples at 0.3 threshold. * indicates that the method might benefit from taking the best matching pairs (see Section~\ref{sec:ssl_evaluation}). }
\label{tab:ssl-music}
\end{table*}

We evaluate our model on single- and multi-source sound localization for scenes containing musical instruments and human speech.

\subsection{Implementation}

\paragraph{Image encoder.} 
We use ResNet-18~\cite{he2016deep} as the backbone for the image encoder. During the training, each frame is randomly cropped and resized to $224 \times 224$. During inference, we directly resize the images without cropping. We encode the image such that the feature map will be down-sampled to $\frac{W}{16} \times \frac{H}{16} \times C$  dimensional embedding vectors. We $l_2$ normalize them along the channel axis, following~\cite{wu2018unsupervised}. During testing, for the synthetic VGGSound and VoxCeleb2 datasets, we concatenate two images so that input of the image encoder is $448 \times 224$ and the output score map is $28 \times 14$. This will keep the aspect ratio of images similar during training and testing. For all other experiments, we use $224 \times 224$ images and $14 \times14$ score maps. We apply bilinear interpolation to upsample score maps for all the methods. 

\mypar{Audio encoder.} 
We use a ResNet-18 network~\cite{he2016deep} to extract $k$ different $l_2$-normalized $C$-dimensional feature vectors from a 0.96s of sound, using a log-mel spectrogram input representation. We compute different embedding vectors for audio nodes by applying different fully connected layers to the final pooled convolutional features. 
We use dot product between image and audio features to calculate the similarity score between image regions and audio nodes. 

\mypar{Hyperparameters.} 
During training, we use the Adam optimizer~\cite{kingma2014adam} with a learning rate of $10^{-4}$ on MUSIC and VGGSound dataset, and a learning rate of $10^{-5}$ on VoxCeleb dataset. We use a batch size of 128 and set the temperature $\tau = 0.07$ following~\cite{wu2018unsupervised}. We set the feature dimension $C = 128$. When processing the audio, the sounds are resampled to 16kHz. The 0.96s audio clips are converted to mel spectrograms of size $ 193 \times 64$ via the Short-Time Fourier Transform (STFT) using 64 mel filter banks, a window size of 160, and a hop length of 80. %

\subsection{Dataset}

\vspace{3mm}

\mypar{MUSIC.}
The MUSIC dataset~\cite{zhao2018sound} contains 11 instruments, in both solos and duets. We use the same training/testing splits as in Hu~\etal~\cite{hu2020discriminative}. The MUSIC-Synthetic dataset~\cite{hu2020discriminative} contains concatenated images with frames from four videos, and the audio is synthesized such that there are two instruments making sounds while the other two are silent. The MUSIC-Duet dataset is a subset of MUSIC that contains duets videos with two instruments playing sounds. We use the solo videos, MUSIC-Solo, to evaluate the performance of single sound source localization, and use MUSIC-Duet and MUSIC-Synthetic datasets when evaluating multiple sound source localization, using the same annotations as~\cite{hu2020discriminative}. %

\mypar{VGGSound-Instruments.} We also evaluate on VGGSound~\cite{chen2020vggsound}. Each video in VGGSound only has a single category label. Analogous to~\cite{arandjelovic2018objects}, we filtered and sampled 37 video classes of musical instruments with 32k video clips of 10s length, and we call this subset VGGSound-Instruments.
The category list is provided in the supplementary material. For evaluation, we filtered and annotated segmentation masks for 446 high-quality video frames\footnote{These annotations can be found at \url{https://web.eecs.umich.edu/~ahowens/mix-localize/}}. When evaluating multi-source localization, we randomly concatenate two frames, resulting in $448 \times 224$ input images, and obtain sound mixtures by summing their waveforms.

\mypar{Human speech.} 
VoxCeleb2 dataset~\cite{Chung18b} is a large-scale audio-visual speaker recognition dataset containing over 1.5k hours of video for 6,112 celebrities. For evaluation, we use a face detector to annotate face segmentation masks for 1k random samples in the test set. We follow the same strategy as in VGGSound-Instruments to create a synthetic multi-speaker synthetic evaluation set.

\subsection{Evaluated methods}

We compare our model with several other audio-visual learning methods. When re-implementing, we use ResNet-18 as our backbone architecture to ensure fair comparisons. 
\mypar{Self-supervised methods.} We compare our model with several variants of the model from Arandjelovi\'{c} and Zisserman~\cite{arandjelovic2018objects}, which we call \textit{OTS}. We follow the model architecture of~\cite{hu2020discriminative} to implement the methods, which uses ResNet-18 to extract the features and global max pooling layer for the fused attention map.
We keep the data prepossessing and network architectures the same as our method when re-implementing them.
We also created a variation of OTS that is trained on synthetic mixtures and concatenated video frames, following \cite{hu2020discriminative}. We call this method \textit{OTS-mix}. 
In contrast to approaches that use multiple frames from videos~\cite{afouras2018conversation, afouras2020self, rouditchenko2019self, zhao2018sound} or require extra manual labels~\cite{hu2020discriminative, qian2020multiple}, these two methods only use one frame and are trained in a fully self-supervised manner.

\mypar{Semi-supervised methods.}
We also consider the semi-supervised, multi-source methods proposed for musical instrument localization in Hu~\etal~\cite{hu2020discriminative}. At training time, these methods cluster the features of training data and match the clusters with ground-truth labels. The sounding and silent visual areas can be obtained by retrieving similarity maps corresponding to the clusters. Since this method uses ground-truth labels to match the clustering result with the classes, we consider it to be semi-supervised. Additionally, we compare with a variant of Sound of Pixels~\cite{zhao2018sound} from \cite{hu2020discriminative}, in which the model predicts 11 different score maps, and uses the training set labels to match the predictions with different classes. We call this method \textit{Sound of Pixels + Matching}.

\begin{table}[!t]
\renewcommand\arraystretch{1.2}
\centering
\resizebox{1\columnwidth}{!}{
\begin{tabular}{clccccccc}

\toprule
&  & \multicolumn{3}{c}{\textbf{Single source}} & \multicolumn{4}{c}{\textbf{Multiple sources}} \\

\cmidrule(lr){3-5}
\cmidrule(lr){6-9}
 & & AP & AUC  & IoU@0.3  & CAP & PIAP & AUC & CIoU@0.1 \\
\midrule
\parbox[c]{2mm}{\multirow{3}{*}{\rotatebox[origin=c]{90}{ {VGGSound}}}} &
OTS~\cite{arandjelovic2018objects} & \textbf{47.3} & 24.5 & 25.7 & \textbf{23.2} & \textbf{37.6} & 10.8  & 51.1  \\

& OTS-mix~\cite{arandjelovic2018objects} & 37.0 & 20.9 & 24.9 & 18.1 & 30.7 & 10.8 & 50.7 \\   
\cdashlinelr{2-9} 
& Ours   & 44.7 & \textbf{32.1} & \textbf{49.6} & 21.5* &  37.4 & \textbf{15.5}* & \textbf{73.1}*\\
\midrule
\parbox[c]{2mm}{\multirow{3}{*}{\rotatebox[origin=c]{90}{ {~VoxCeleb2~}}}} &
OTS~\cite{arandjelovic2018objects} & 43.9 & 23.5 & 6.2  & \textbf{20.4}  & 32.6 & 7.0 & 15.8  \\

& OTS-mix~\cite{arandjelovic2018objects}  & 21.4 & 6.4 & 6.2 & 10.7 & 18.2 & 4.1 & 15.8 \\   
\cdashlinelr{2-9} 
& Ours   & \textbf{46.1} & \textbf{27.7} & \textbf{35.4} & 20.1* & \textbf{35.4} & \textbf{14.2}* & \textbf{17.4}*\\

\bottomrule
\end{tabular}
}
\caption{Sound source localization performance on VGGSound-Instruments and VoxCeleb2 datasets. \vspace{-0.3em}}
\label{tab:ssl-vggsound}

\end{table}

\begin{figure*}[t]
    \centering
    \includegraphics[width=1\textwidth]{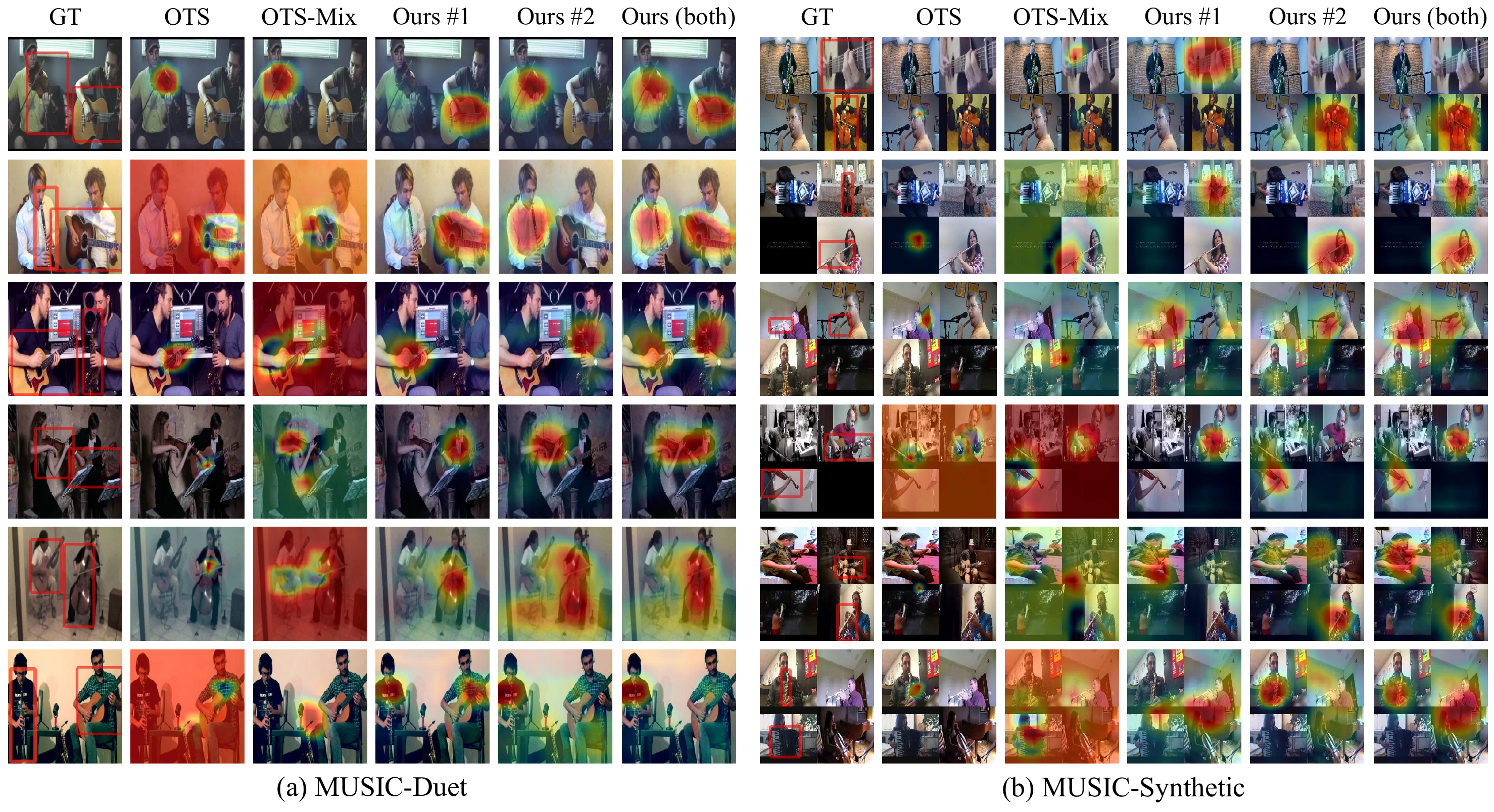}
    \vspace{-5mm}
    \caption{
    {Multi-source sound localization on the MUSIC dataset~\cite{zhao2018sound,zhao2019sound}. We show failure cases in the last two rows.%
    }}
    \label{fig:ssl}
\end{figure*}

\subsection{Evaluation of sound source localization}
\label{sec:ssl_evaluation}
We evaluate these methods on both single- and multi-source sound localization. Unlike methods that produce only one localization map for all the sounding objects in the scene, our method localizes multiple objects (by producing $k$ localization maps). We therefore expect our method to perform approximately as well when compared with other methods on single sound-source localization, and to outperform these methods on multiple sound localization. 

\mypar{Evaluation of single sound source localization.}
We evaluate the single sound source localization performance following~\cite{hu2020discriminative, senocak2018learning}. Given the ground truth bounding boxes or object segmentation mask, we compute the \textit{Intersection over Union} (IoU) and \textit{Area Under Curve} (AUC) according to the predicted sounding area. For methods such as Objects that Sound~\cite{arandjelovic2018objects} (OTS) that produce one output, we take the produced single sounding area to compute the score. For our method, since there are $2$ sounding area maps (corresponding to 2 audio nodes), we take the average of these maps as the final sounding area. When computing the scores, we use a threshold of 0.4 for our method in all experiments. For other methods, we choose the best threshold for each method according to the performance on validation set. Additionally, to avoid judging the methods based on a fixed threshold, we also use pixel-wise average precision (\textbf{AP})~\cite{choe2020evaluating}.

\mypar{Evaluation of multiple sound source localization.} 
We follow~\cite{hu2020discriminative} and use Class-aware IoU (CIoU) when evaluating multiple sound source localization. Analogously, we propose to use class-aware average precision (CAP) to provide a threshold-less evaluation metric. The CAP score is calculated as:
\begin{equation}
CAP = \frac{\sum_{k=1}^{K} \delta_k AP_{k}}{\sum_{k=1}^{K} \delta_k},
  \label{eq:CAP}
\end{equation}
where $AP_{k}$ measures the pixel-wise average precision for the class $k$. The indicator $\delta_k$ indicates whether an object of class $k$ is making sound. Since our method does not have class labels (\ie, we do not know which localization map corresponds to which object), we use a modified version of CAP for our method, where we evaluate both pairings of the predicted and ground-truth labels, and report the best. %
Since this provides a potential unfair advantage to our method, we also introduce another metric called permutation-invariant average precision~(PIAP). When computing this score, we take the average of the sounding area maps and compute the average precision using the ground truths of all sounding objects. 

\subsubsection{Single-source localization}
\label{single-source-localization}
We evaluate the performance of single sound source localization on MUSIC-Solo, VGGSound-Instruments and VoxCeleb2. Under this evaluation setup, the input audio to the models is original unmixed audio instead of mixture sounds. The results are shown in Table~\ref{tab:ssl-music} and Table~\ref{tab:ssl-vggsound}. It can be seen that our method performs approximately equally well on single sound source localization when compared with other methods. This suggests that our method is capable of localizing a single sound source. 
We find that when the input audio is from a single source (i.e., unmixed), the model tends to predict two similar localization maps. We note that these single-source sounds were not explicitly provided during training.

\subsubsection{Multi-source localization}
\vspace{3mm}
\mypar{Quantitative results.}
We evaluate the multiple sound source localization performance on MUSIC-Duet, MUSIC-Synthetic, VGGSound-Instruments and VoxCeleb2 datasets in Table~\ref{tab:ssl-music} and Table~\ref{tab:ssl-vggsound}. The comparison with other work shows that our proposed method achieves better performance on the multiple sound source localization task. We note that our method does not use labels (unlike~\cite{hu2020discriminative}), or multiple frames (unlike~\cite{zhao2018sound}). Instead of taking synthetic data as input as~\cite{hu2020discriminative}, we use the unmodified images in the dataset, which might be the reason why our method performs better on  MUSIC-Duet than MUSIC-Synthetic. The experiment results indicate that the proposed cycle-consistency approach leads to improvements in multi-source sound localization.

\mypar{Qualitative results.}
In Figure~\ref{fig:vggsound_ssl} and Figure~\ref{fig:ssl}, we visualize the localization maps generated by these methods. 
It can be seen that models based on Objects that Sound (OTS)~\cite{arandjelovic2018objects} mainly focuses on one of the sounding objects, rather than all sounding objects, while our method spreads the probability to all objects. In particular, the audio features obtained by our approach group the visual regions corresponding to both sound sources. For the qualitative results on VoxCeleb2, we found that the model fails more often when the gender of the two speakers is the same. %

\begin{table}[!t]
\centering
\resizebox{1\columnwidth}{!}{
\begin{tabular}{lcccccccc}

\toprule
& \multicolumn{4}{c}{MUSIC-Synthetic} & \multicolumn{4}{c}{MUSIC-Duet} \\
\cmidrule(lr){2-5} 
\cmidrule(lr){6-9} 
 & CAP & PIAP & \hspace{0.0em}AUC\hspace{0.0em}  & CIoU@0.3  & CAP & PIAP & \hspace{0.1em}AUC\hspace{0.0em}  & CIoU@0.3   \\
\midrule

Corre  & 12.6 & 16.0 & 7.3 & 0.0 & 19.3 & 21.1 & 17.8 & 7.7 \\   
ISI            & 11.0 & 16.4 & 7.3 & 0.0  & 19.7 & 24.9 & 17.5 & 7.6  \\
Permute     & 18.2* & 24.4 & 9.1* & 0.4* & 24.0* & 28.1 & 19.5* & 12.4* \\ 
\cdashlinelr{1-9} 
Ours         & \textbf{34.0*} & \textbf{39.7} & \textbf{20.2*} & \textbf{21.3*} & \textbf{47.4*} & \textbf{53.9} & \textbf{21.2*} & \textbf{26.3*} \\

\bottomrule
\end{tabular}
}
\caption{\textbf{Ablation study.} We evaluate the sound source localization performance on MUSIC-Synthetic and MUSIC-Duet datasets for each ablation model. \textit{Corre} denotes the mixed correspondence model while \textit{Permute} represents the model with permutation invariant loss.\vspace{-0.5em} }
\label{tab:ablation}
\end{table}

\subsection{Ablation study}
We also explore a number of variants of our model for training the self-supervised audio-visual system. We compare our model with several other designs. We keep all the settings the same except the loss function. 

\mypar{Image-audio-image cycle.}
We consider cycles that start from image nodes, rather than audio nodes. This walk starts at an image, goes to the audio nodes, and finally cycles back to the image nodes. The two image nodes here are sampled from the same video, such that the semantic meaning of the nodes does not change. The similarity between two images, therefore, is evaluated by the probability that they reach each other on a cross-modal random walk. This model minimizes the loss:
\begin{equation}
\begin{aligned}
     \mcl_{\mathtt{ISI}} = \frac{1}{k}\tr(\log({A}_{IS} A_{SI})).
  \label{eq:loss_cyc_2}
\end{aligned}
\end{equation}
\noindent
We call this model the {\em ISI} model, due to the image-sound-image path that the random walker takes.

\mypar{Mixed correspondence loss.} To test whether the model benefits from cycle-based training (rather than other model differences), we compared with a model trained with the InfoNCE~\cite{oord2018representation} loss with exactly the same input (a single frame and mixed audio). Since we do not know the association between audio nodes and image nodes, we modify Eq.~\eqref{eq:AIS} to account for it, \ie, 
\begin{equation}
\begin{aligned}
{A}_{IS}(i, j) & = \frac{\sum_{t=1}^k \exp(\simfn(\bv_i, \bs_t^{(j)})/ \tau)}{\sum_{\bs_t \in S} \exp(\simfn(\bv_i, \bs_t)/ \tau)}, 
  \label{eq:corresp}
\end{aligned}
\end{equation}
where $\bs_t^{(j)}$  is one of the $k$ audio embeddings generated by the mixed audio for example $j$, $S$ is the set of all audio embeddings in the batch, and $\simfn$ is defined as in Eq.~\eqref{eq:AIS}. In contrast to our method, this loss obtains significantly more negative samples by obtaining them from other examples in the batch. %

\mypar{Permutation invariant loss.} Inspired by audio source separation methods~\cite{wisdom2020unsupervised,hershey2016deep}, we ask whether the association between audio and images can be learned from a {\em permutation invariant} loss. We consider all possible pairings of image and audio embeddings, and select the one with the largest total similarity. For $k=2$, this loss is:
\begin{equation}
\begin{aligned}
\mcl_{\mathtt{PIT}} = %
-\max(&\simfn(\bv_i, \bs_1) + \simfn(\bv_j, \bs_2), \\ & \simfn(\bv_i, \bs_2) + \simfn(\bv_j, \bs_1)),  
  \label{eq:pit}
\end{aligned}
\end{equation}

\noindent
where $\bv_i$ and $\bv_j$ are a pair of images used to create a synthetic mixture, and $\bs_1$ and $\bs_2$ are the audio embeddings produced from their mixed sound. While this loss is similar to $\mcl_{\mathtt{cyc}}$ (Eq.~\ref{eq:loss_cyc}), it creates a ``hard" association between images and audio embeddings via the $\max$ operation. By contrast, the random walk model makes ``soft" assignments between the embeddings that during learning provides a gradient signal for both possible pairings.

\begin{figure}[t]
    \centering
    \includegraphics[width=1\columnwidth]{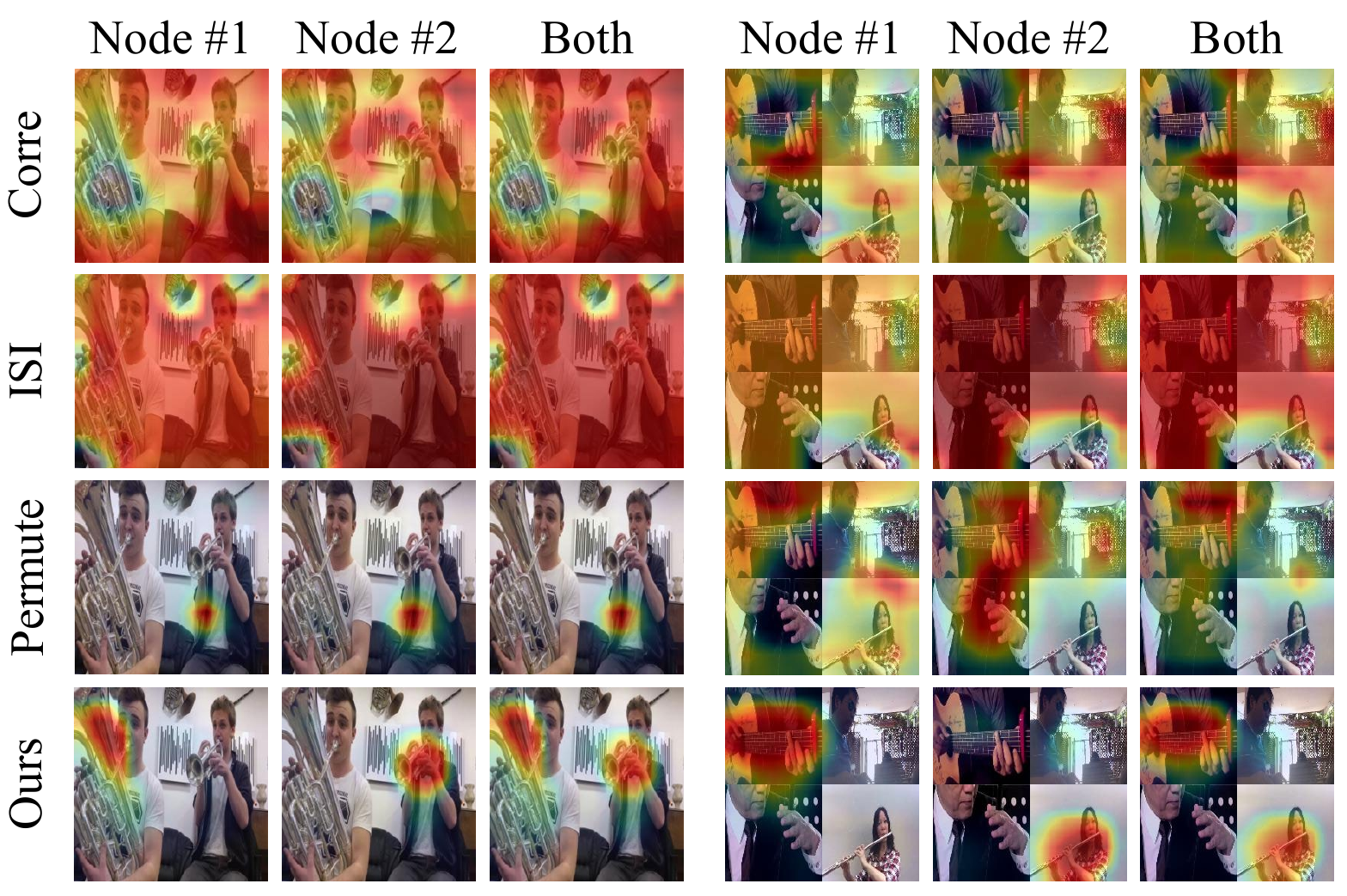}
    \caption{{\textbf{Ablation study.} We visualize the score maps for these methods on MUSIC-Synthetic and MUSIC-Duet dataset. } \textit{Corre} denotes the mixed correspondence model while \textit{Permute} represents the model with permutation invariant loss.}
    \label{fig:ablation}
\end{figure}

\mypar{Results.}
We evaluated multi-source localization on MUSIC-Synthetic and MUSIC-Duet datasets in Table~\ref{tab:ablation}. In Figure~\ref{fig:ablation}, we visualize the score maps predicted by these methods. These methods fail to produce different and correct localization maps for the two audio nodes, indicating they fail to produce different audio embeddings for each sound source. It can be seen that our method outperforms all these variants. 

Although the ISI model is also based on cycle consistency, it does not  learn to explicitly separate the two audio nodes. By contrast, our model needs to create two distinct embeddings for the audio nodes in order to successfully complete a cycle. Compared to mixed correspondence loss, which uses other images and audio in the batch as the negative samples for contrastive learning, our method instead takes advantage of other audio nodes derived from the same mixed audio. This will also encourage the model to separate different audio nodes. Moreover, unlike the permutation invariant model that requires a ``hard" correspondence for every pair of images and audio, our method allows the model to assign a probability to the graph edge.

\section{Discussion}
In this paper, we have proposed a simple, self-supervised method for visually localizing sounds in audio mixtures. Our approach is based on learning a cycle-consistent random walk on a graph that connects nodes defined by the images and sound. %
We showed that our method identifies and segments multiple sound sources more accurately than other self-supervised methods based on traditional audio-visual correspondence learning. 

Our results suggest that cycle-consistent random walks can be used to successfully group the contents of a multimodal scene into distinct objects. We hope that these techniques can be combined with tracking-based cycle consistency~\cite{wang2019learning,jabri2020space} to group the scene contents over time as well. 
We also hope this approach provides further directions for the study of cross-modal learning. One such direction is to directly combine explicit source separation~\cite{hershey2016deep} with the ``implicit" source separation we obtain through contrastive learning.

\mypar{Limitations.} 
Our released models are limited in scope to the benchmark video datasets they were trained on. Since these are popular datasets, information about their biases is publicly available. As in other audio-visual speech work~\cite{nagrani2018seeing} the learned models may learn to correlate visual properties of a speaker with their voice, making them susceptible to bias. %

\mypar{Acknowledgements.}
We thank Di Hu and Yake Wei for the help with the experimental setup. We would also like to thank Linyi Jin for suggestions on figure design.
This work was funded in part by DARPA Semafor and Cisco Systems. The views, opinions and/or findings expressed are those of the authors and should not be interpreted as representing the official views or policies of the Department of Defense or the U.S. Government.

{\small
\bibliographystyle{ieee_fullname}
\bibliography{mix}
}

\clearpage
\appendix

\section{Additional qualitative results} 

\begin{figure*}[!htbp]
    \centering
    \vspace{-1em}
    \includegraphics[width=0.95\textwidth]{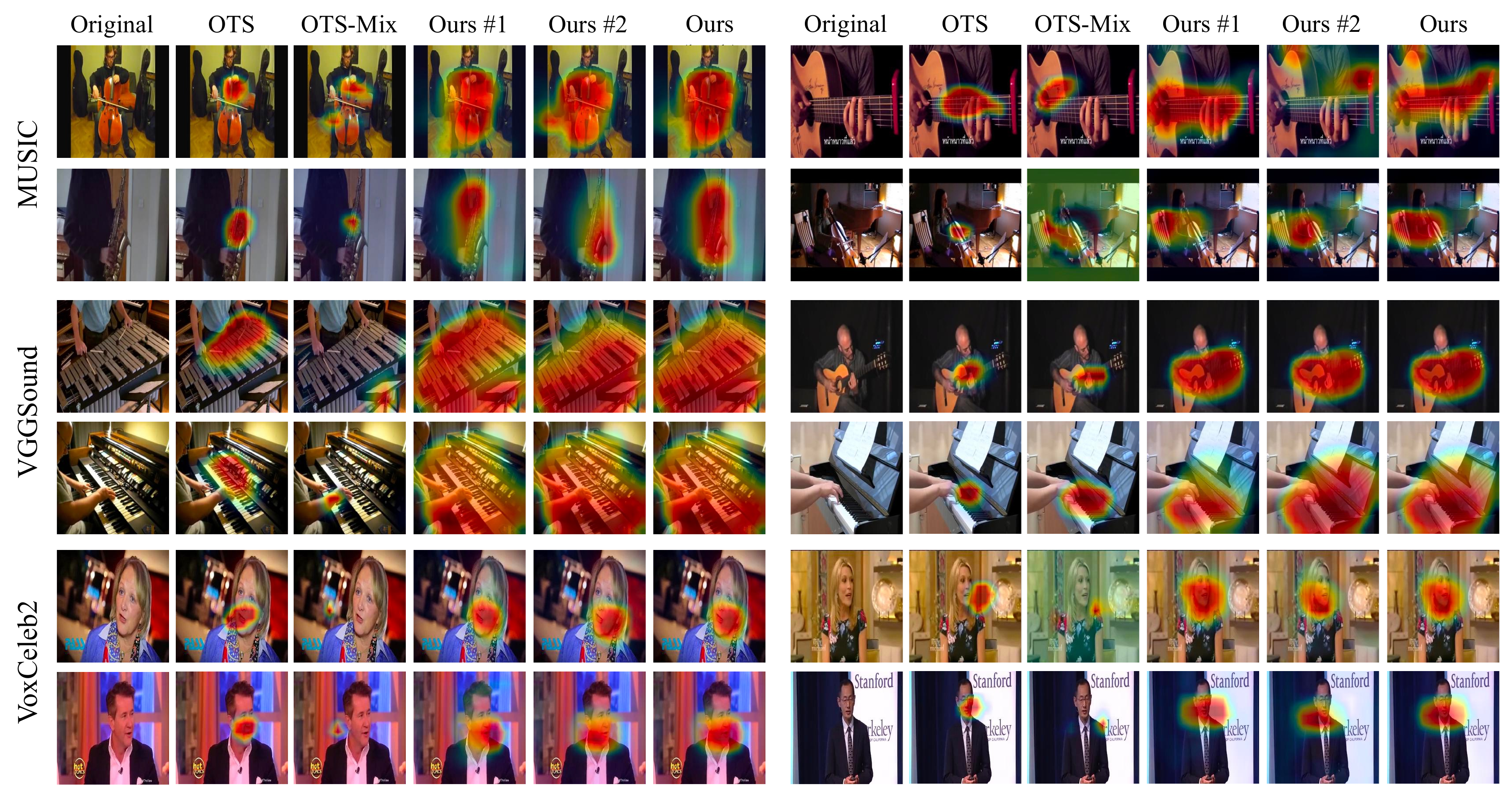}
    \caption{\textbf{Single sound source localization results on VGGSound-Instruments and MUSIC datasets.} Comparison of score maps generated by different methods. \#1 denotes the localization map generated using the first node, and \#2 represents the one generated using the second node. The color of the patch indicates its localization score (red means high, and blue means low). The results in the last three rows are randomly sampled.
    }
    \label{fig:supp-fig-single-ssl}
\end{figure*}

\vspace{2mm}
\mypar{Single sounding object localization.}
We provide the visualization for single sound source localization, as an extension of Section~\ref{single-source-localization}. The results are shown in Fig.~\ref{fig:supp-fig-single-ssl}. We can see that our model learns the same localization map for both audio nodes when there's only one sounding object.

\begin{figure*}[!htbp]
    \centering
    \vspace{-1em}
    \includegraphics[width=1\textwidth]{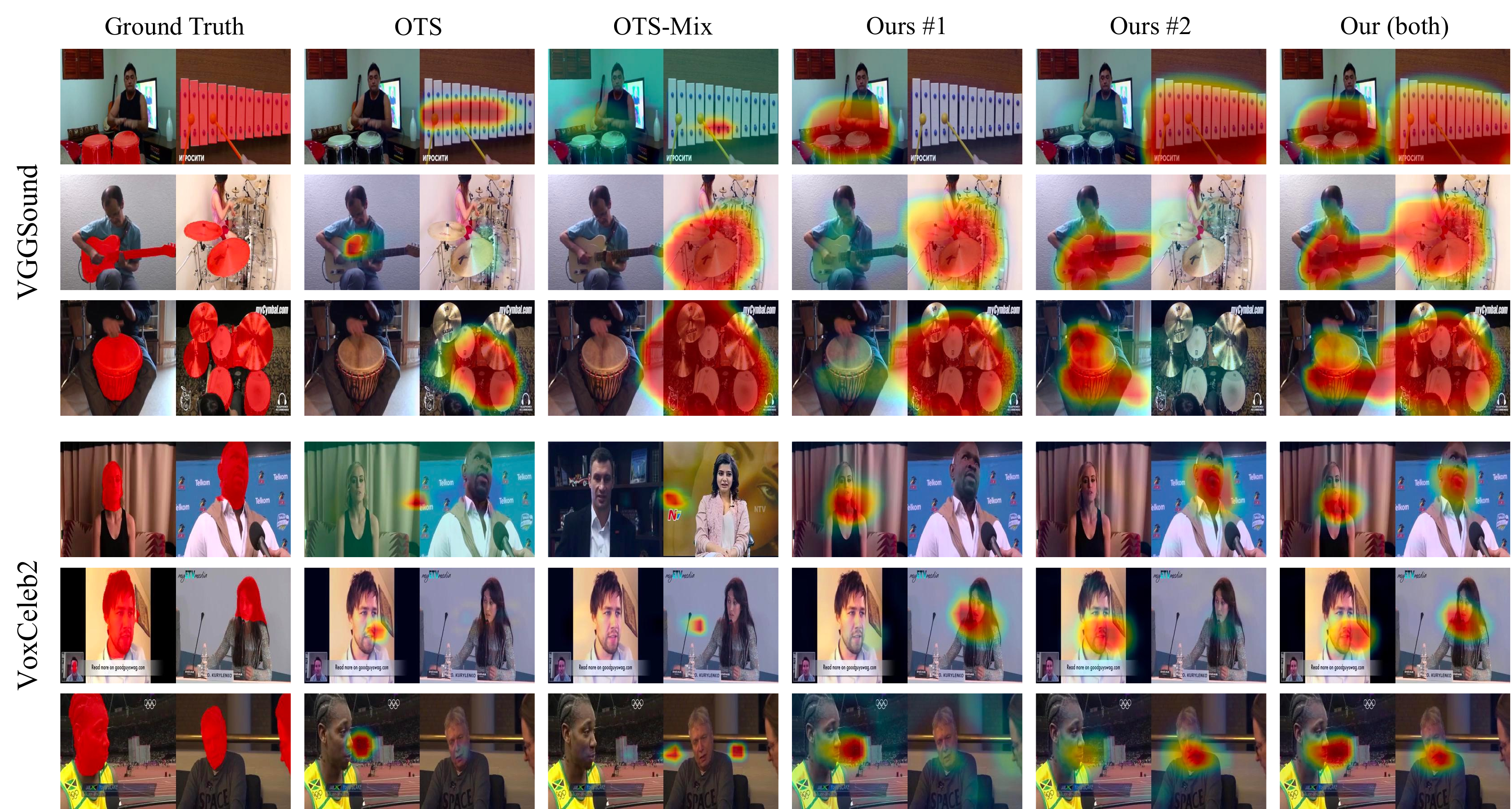}
    \caption{\textbf{Multiple sound source localization results on VGGSound-Instruments and VoxCeleb2 dataset.} Comparison of score maps generated by different methods. \#1 denotes the localization map generated using the first node, and \#2 represents the one generated using the second node. The color of the patch indicates its localization score (red means high, and blue means low). The results are randomly sampled. }
    \label{fig:supp-fig-vggsound-multi-ssl}
\end{figure*}

\mypar{Multiple sounding object localization. }
We show more qualitative results of multiple sounding object localization in Fig.~\ref{fig:supp-fig-vggsound-multi-ssl} and Fig.~\ref{fig:supp-fig-music-syn-ssl}, which are the extensions of Fig.~\ref{fig:vggsound_ssl} and Fig.~\ref{fig:ssl} in our paper. We also include the model of Chen \etal~\cite{chen2021localizing}, using their publicly available VGGSound model. It can be seen that our method with cycle loss outperforms those learning solely from audio-visual correspondence. 

\section{Visualization of learned cycle paths.} 
We provide the visualizations of the random walk probabilities for several examples. 
\begin{figure}[H]
    \centering
    \includegraphics[width=1.0\columnwidth]{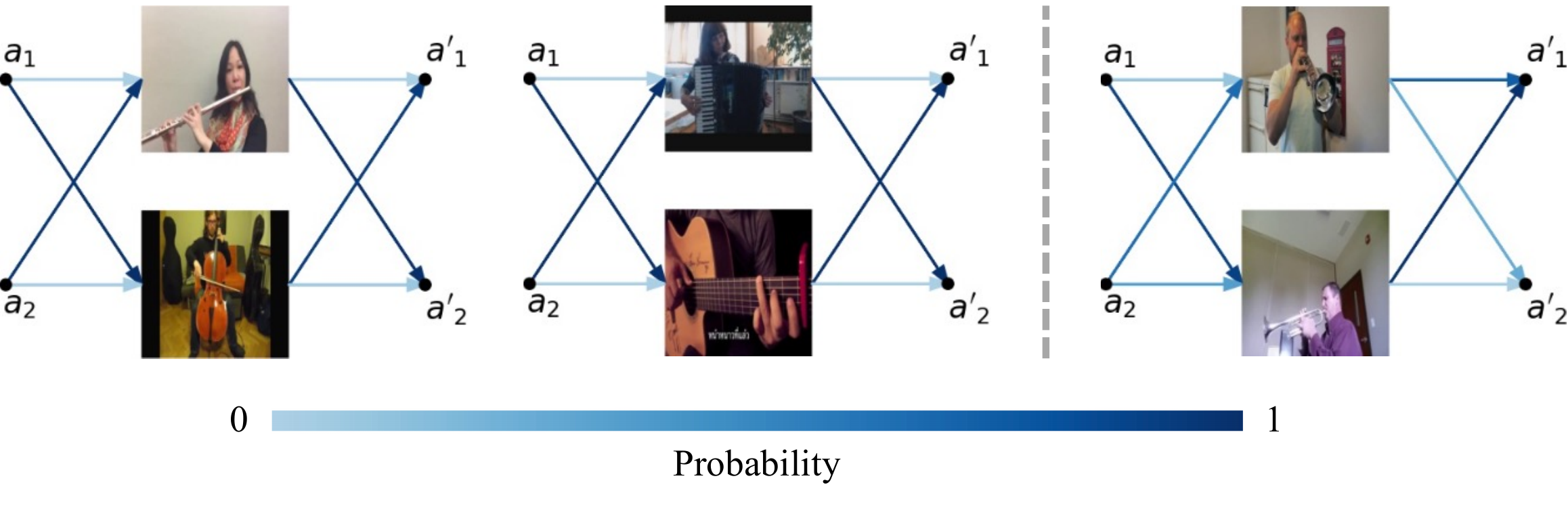}
    \vspace{-2em}
    \caption{Visualization of cycle paths. The color of the edge connecting two nodes indicates its transition probability. The last sample is an ambiguous mixture of two accordions.}
    \label{fig:rebuttal-cycle-paths}
\end{figure}

\begin{minipage}{\textwidth}
\centering
    \includegraphics[width=1.0\textwidth]{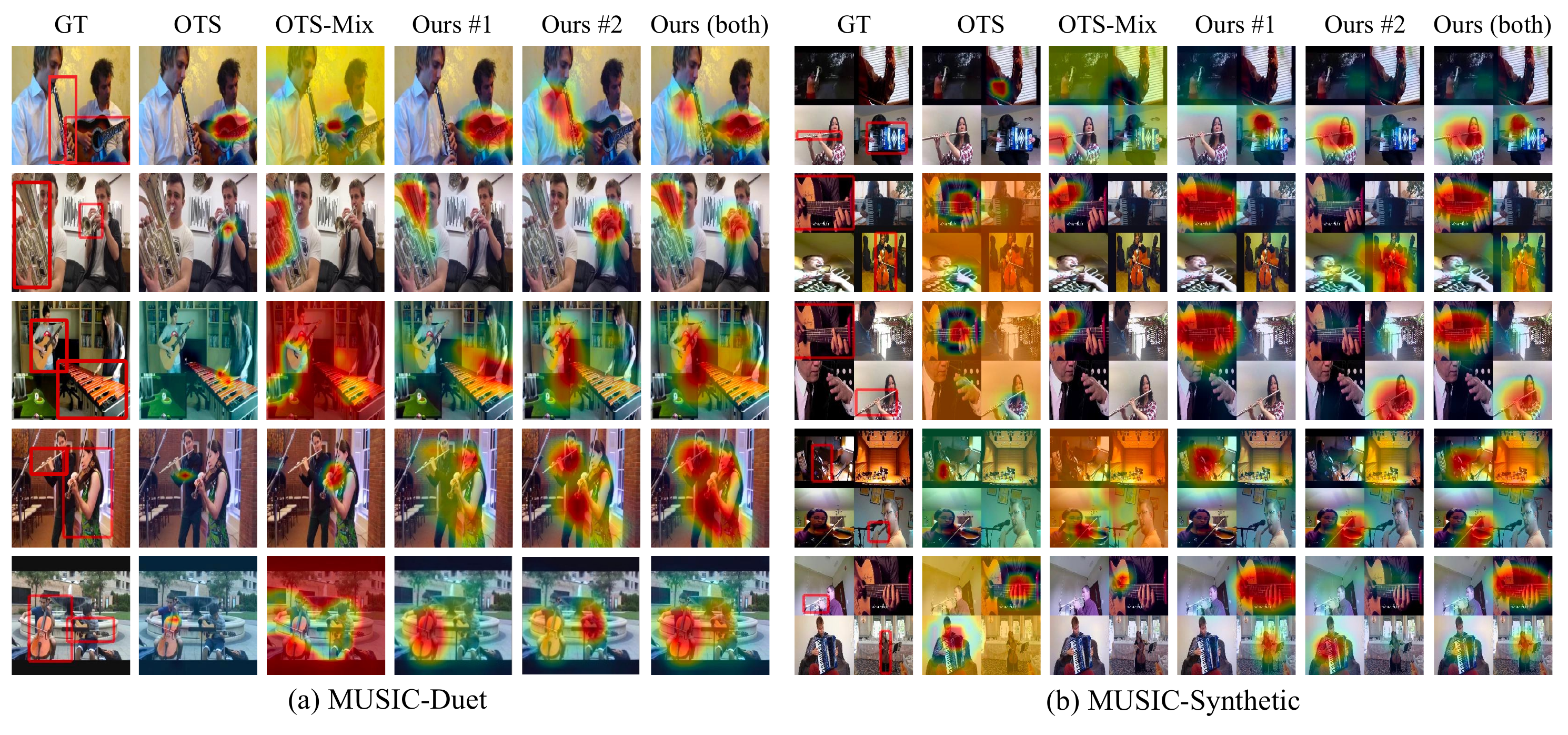}
    \vspace{-5mm}
    \captionof{figure}{\textbf{Multiple sound source localization results on MUSIC-Synthetic dataset.} Comparison of score maps generated by different methods. \#1 denotes the localization map generated using the first node, and \#2 denotes the one generated using the second node. The color of the patch indicates its localization score~(red means high, and blue means low).}
     \label{fig:supp-fig-music-syn-ssl}
\end{minipage}

\section{Category list for VGGSound-Instruments.}
For the VGGSound dataset, we sampled 37 video classes of musical instruments with 32k video clips of 10s length, and we call this subset VGGSound-Instruments. Video classes are listed below: 
\vspace{-0.5em}
\begin{table}[h]
\begin{center}
\resizebox{1\columnwidth}{!}{
\begin{tabular}{l l l }
playing accordion       &   playing acoustic guitar   &
playing banjo           \\   playing bass drum   &
playing bass guitar      &
playing bongo        \\    playing cello           &   
playing clarinet        &   playing congas  \\
playing cornet  &
playing cymbal          &   playing djembe   \\
playing double bass     &   playing drum kit   &
playing electric guitar \\ 
playing electronic organ &  playing erhu    &
playing flute   \\ playing glockenspiel   &
playing guiro  &
playing hammond organ   \\   playing harp   &
playing harpsichord     &   playing mandolin   \\ 
playing marimba, xylophone   &
playing piano   &  playing saxophone       \\   
playing sitar   &   playing snare drum      &   
playing steel guitar, slide guitar \\
playing tabla           &   playing timbales  &
playing trumpet         \\   playing ukulele &
playing vibraphone      & 
playing violin, fiddle   \\  playing zither          &    &        \\
\end{tabular}
}
\end{center}
\label{tab:category}
\end{table}

\vspace{-2em}

\end{document}